\documentclass[11pt,a4paper]{article}
\usepackage[hyperref]{ranlp2021}
\usepackage{times}
\usepackage{latexsym}

\usepackage{microtype}

\usepackage{amsmath,amssymb,amsfonts}
\usepackage{lipsum}% http://ctan.org/pkg/lipsum
\usepackage{algorithmic}
\usepackage{graphicx}
\usepackage{enumitem}
\usepackage{textcomp}
\usepackage{multirow}
\usepackage{xcolor}

\aclfinalcopy

\title{Classification of Consumer Belief Statements From Social Media}

\author{
Gerhard Hagerer\textsuperscript{{\normalfont 1}} \and 
Wenbin Le\textsuperscript{{\normalfont 1}} \and 
Hannah Danner\textsuperscript{ {\normalfont 2}} \and
Georg Groh\textsuperscript{{\normalfont 1}}  \\
\textsuperscript{1}Social Computing Research Group, Technical University of Munich\\
\textsuperscript{2}Chair of Marketing and Consumer Research, Technical University of Munich\\
\texttt{\{ghagerer,grohg\}@mytum.de}
}

\date{}

\begin{document}
\maketitle

\begin{abstract}
Social media offer plenty of information to perform market research in order to meet the requirements of customers. One way how this research is conducted is that a domain expert gathers and categorizes user-generated content into a complex and fine-grained class structure. In many of such cases, little data meets complex annotations. It is not yet fully understood how this can be leveraged successfully for classification. We examine the classification accuracy of expert labels when used with a) many fine-grained classes and b) few abstract classes. For scenario b) we compare abstract class labels given by the domain expert as baseline and by automatic hierarchical clustering. We compare this to another baseline where the entire class structure is given by a completely unsupervised clustering approach. By doing so, this work can serve as an example of how complex expert annotations are potentially beneficial and can be utilized in the most optimal way for opinion mining in highly specific domains. By exploring across a range of techniques and experiments, we find that automated class abstraction approaches in particular the unsupervised approach performs remarkably well against domain expert baseline on text classification tasks. This has the potential to inspire opinion mining applications in order to support market researchers in practice and to inspire fine-grained automated content analysis on a large scale.
\end{abstract}

\section{Introduction} \label{introduction}

The rise of social media has enabled and sparked user's desires to share opinions publicly online. The user-generated content often reveals their true \textit{customer beliefs} towards a certain aspect of things and therefore worth being researched. Typical research fields are qualitative social studies, e.g., \textit{content analyses} for market and consumer research as well as political surveys. One of the challenges, however, has been the course of parsing and organizing a large amount of data that is becoming available in the form of natural language into a fine-grained class structure, which provides a more digestible and actionable insight. This can be achieved by injecting the knowledge from domain experts through annotations. In that regard, there is much manual labour and massive associated expenses invested for such content analysis on social media texts.

In the corresponding qualitative research, fine-grained expert annotations are provided, which have a high resolution on few data points compared to what is available within social media. This is contrary to the requirements of automated textual analyses, e.g., \textit{supervised classification} based on natural language processing and machine learning. Consequently, the question if and how that type of \textit{domain-specific, expert-annotated data} can be effectively leveraged for automated large-scale analyses is a challenging research problem. It carries high potential to gain insights into data which goes beyond a few manually selected texts, scaling up the derived opinion mining models to gather relevant statistics over big amounts of related textual social media corpora, i.e., to inform automated opinion mining based analysis.

%One requirement of automated textual analyses is to provide a sufficient number of data points for each class to be able to train classification algorithms. 

For their own qualitative content analysis, domain experts provide fine-grained labels. These, in turn, are organized in a meaningful hierarchy, such that coarse-grained classes of texts always contain several fine-grained classes. For example, a class of social media comments about food products might contain statements including fine-grained attributes, such as, \textit{taste}, \textit{safety}, \textit{price}, etc. \textit{Safety} could in turn contain even more fine-grained \textit{belief statements} about \textit{food products containing chemicals}, \textit{safe and regulated foods}, \textit{unsafe nutrition}, and so on. This shows there are labels at different levels inside of the class hierarchy of customer belief statements. 

As we are interested in optimizing the conditions towards improved supervised classification performance, we see the potential to compare supervised classification at different levels of such a class hierarchy, as higher, more abstract classes contain more datapoints and thus solidify model training. We investigate if the expert-based class hierarchy is optimal for classification purposes, or if automatic, hierarchical clustering of the classes yields labels, which would be more suitable for supervised classification. This could notably reduce human effort as well as labour cost and increase productivity to thus accelerate the research process. Automated approaches also provide consistent results with less variability than humans. Not to mention, how it could reactivate existing meaningful content analyses to provide new insights at little cost. We perceive a great potential in semi-automated class abstraction, especially for large-scale content analysis.
%We investigate different class abstraction approaches on the organic dataset for text classification tasks. Class abstraction is achieved by combining many fine-grained classes with few samples to few coarse-grained classes with many samples. This can be achieved manually by a domain expert or automatically by hierarchical clustering based on given fine-grained classes. The generation of fine-grained classes in the first place can also be achieved manually by a domain expert or by unsupervised techniques such as traditional clustering methods, e.g., k-means. 
More precisely, we are investigate the following research questions (RQs):

\def \rqone{How suitable are expert annotations from domain-specific content studies about customer beliefs suitable for automatic classification?}
\def \rqtwo{How does the combination of fine-grained classes to more coarse classes relate to classification accuracy?}
\def \rqthree{Can an automatic combination of fine-grained classes improve classification over manual, expert-based class hierarchies?}
\def \rqfour{How does this compare to classes which are derived without any expert knowledge, i.e., by unsupervised text clustering methods?}
\def \rqfive{What are the favorable and the unfavorable effects of automatic and manual class combination and how do these relate to each other?}
\def \rqsix{How do different pre-trained word and sentence embeddings perform for our objectives?}

\begin{enumerate}[label=RQ\arabic*, leftmargin=8mm]
\setlength\itemsep{0em}
\item \rqone
\item \rqtwo
\item \rqthree
\item \rqfour
\item \rqfive
\end{enumerate}

% To solidify our formulated research questions, we think it is helpful to consider the following technical optimization-related questions in depth:

% \begin{enumerate}[label=RQ\arabic*, leftmargin=8mm]
% \setlength\itemsep{0em}
% \setcounter{enumi}{5}
% \item \rqsix
% %\item \rqseven
% \end{enumerate}

Our research dataset consist of opinions about organic food and related consumer issues on social media in German-speaking countries and the United States, which is described in section 4. We analyze the differences regarding machine learning classification accuracy using labels of varying granularity generated from expert-based and automated class hierarchies as explained in section \ref{sec:methodology}. The latter can be further divided into supervised and unsupervised approaches. In supervised class hierarchies, we form new class hierarchies based on pairwise semantic class similarities and hierarchical clustering of the existing fine-grained expert labels. 
%The degree of class abstraction, i.e., latent space dimensionality, is a parameter we fine-tune in this process. 
This differs from unsupervised class abstraction, where the fine-grained classes are not given by the domain expert but by unsupervised semantic text clustering.
%using transfer learning and clustering techniques. 
We describe how to experimentally compare the regarding classification performances in section \ref{sec:experiments} to see which kind of labeling techniques are beneficial for predictive machine learning models. 
%Regarding the latter, we analyze the difference when using pre-trained word and sentence embeddings in the course of label extraction for unsupervised approach. 
%Also, different document representations are compared as features regarding classification accuracy.
% and reveal the best combination of document representations and class abstraction approaches. 
The results part of section \ref{sec:results} amongst others depicts the effects of class abstraction for classification and according favourable and unfavourable effects.
%as well as how do document representation techniques and pre-trained embeddings affect the classification. 
Section \ref{sec:conclusion} answers the research questions and gives an outlook for future work and potential.

%The contribution of this paper is an analysis of an automated class abstraction approach which creates a meaningful intrinsic class hierarchy. This in turn produces coarse class labels outperforming expert-based class labels in text classification tasks. Section 4 presents first the experiment dataset and explains why a domain-specific one is preferred. Secondly, it introduces the necessary preprocessing steps. Next, we introduce how to utilize directly the knowledge from domain experts as baseline. Most importantly, we present the concepts of supervised and unsupervised class combination approaches with regard to the question how expert labels could be combined into abstract labels based on latent semantic indexing, word mover's distance and bag-of concepts approaches. Besides, we also bring in the knowledge of how it could be done in a completely unsupervised manner using transfer learning and clustering techniques. Finally, we demonstrate the experimental set up for a multi-label classification task including classifier selection, feature choices, and evaluation metrics. Section 5 presents mainly the results. The main contribution is demonstrated as the classification accuracies between class abstraction approaches. 

\section{Related work}

Social media are an established source to investigate consumer beliefs of relevant market domains, for instance, [List some examples]. Therefore, content analyses are carried out, which follow the methodology of \textit{grounded theory} \cite{martin1986grounded}. A domain expert labels the given texts with so called codes, which are tags of ideas or concepts describing the related belief statements. These tend to be rather fine-grained and concrete with a high semantic correspondence to the labeled statements from the text. As an outcome, there are many fine-grained labels with few samples per class, which thus need to be combined to categories. This problem is described as follows:

%In preparing labels for one's concept cards, the initial aim is to find a level of abstraction high enough for one to avoid creating a separate concept card for every "fact" observed but low enough to ensure that the discovered concept relates explicitly to the substantive phenomenon under study. 

\begin{quote}
\textit{``If a label is insufficiently abstract or general, too few observations will fall into that category. One will add few incidents to each concept card as one analyzes the data. In such cases, the label merely restates or rephrases the data. To "work" (Glaser \& Strauss, 1967), a conceptual label must occupy a higher level of abstraction than the incidents (facts, observations) it is intended to classify. If the concept label is too abstract, however, too much information will fall into that category.''} \cite{martin1986grounded}
\end{quote}

To choose a category system is relevant, because this supports the cognitive and scientific progress of formulating theories and gaining insights over the analyzed behavior and attitudes within that domain. Theoretical works find that this type of content analysis is substantially similar to topic modeling, a technique from text mining and natural language processing (NLP) \cite{10.1007/978-3-030-33232-7_25,yu2011compatibility,piepenbrink2017topic}. As a consequence, these text mining techniques are adopted in practice for content analyses on social media, especially for consumer research \cite{rocklage2019text}. It is a useful means to support decision makers for product development by providing them with information about a market segment, competing companies, and consumer requirements \cite{XU2011743,HE2013464}. However, this research does not discuss or leverage NLP methods to detect fine-grained consumer beliefs, which express another dimension of the consumer attitude than the sentiment-related affect \cite{pernerconsumer}. In fact, there is little research on if and how consumer beliefs can be classified by NLP classifiers, as these are primarily concerned with the task of sentiment analysis. However, there is evidence that the task of consumer belief classification can be solved by proper NLP methods, including topic modeling as a means to detect new and yet unknown beliefs. The granularity of these, however, is not examined, not to speak of how this relates to expert-based annotations and judgements.

\begin{figure}[t]
\centerline{\includegraphics[width=0.5\textwidth]{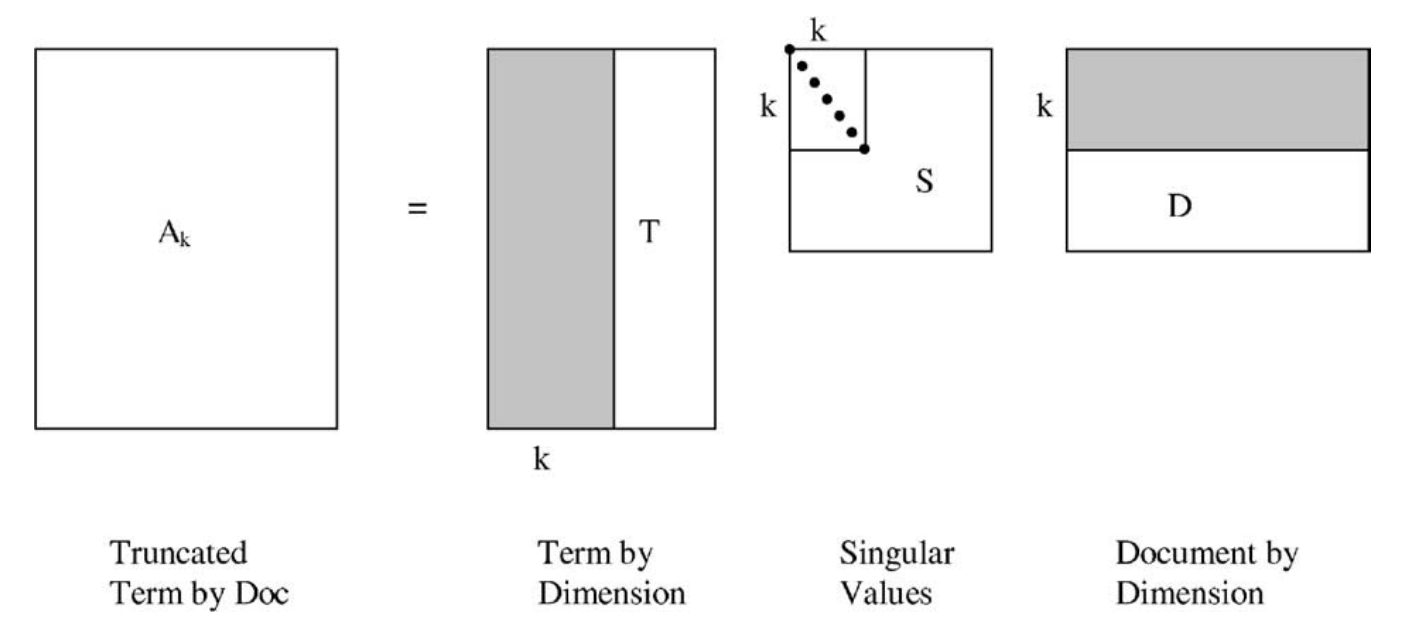}}
\caption{LSI is a SVD decomposing a term-document matrix A into a term-dimension matrix T, a singular-value matrix S, and a document-dimension matrix D, which are all reduced to k latent topics.  \cite{b3}}
\label{LSI}
\end{figure}

\section{Methodology} 
\label{sec:methodology}

\subsection{Hierarchical Class Clustering}

We use agglomerative hierarchical clustering  \cite{b6} to merge our expert classes into automatically generated class hierarchies based on semantic distance metrics. Since hierarchical clustering is a bottom-up approach, each expert label is first considered as a single-element cluster. We use the weighted-average linkage criteria (WPGMA), which calculates the intra-cluster distance intuitively as the arithmetic mean when forming new clusters. 

\paragraph{Semantic Similarity Metric}

For hierarchical clustering of the given expert classes, we use LSI as semantic distance metrics.

%\paragraph{Latent Semantic Indexing}
LSI \cite{b3} utilizes singular value decomposition (SVD) to reveal latent relationships between documents in a corpus. It yields a low-rank approximation for each document and enables to extract document-document semantic similarity in this low-rank document representation. As illustrated in Fig. \ref{LSI}, SVD decomposes a document-term matrix A into term matrix T, singular value matrix S, and document matrix D, where each matrix will be truncated to k dimensions, i.e., k latent topics. We utilized the document matrix D to construct document representation, in which each document is represented as a linear combination of latent topics. The weights of this combination are taken as the vector representation of the document.

\section{Experiments}
\label{sec:experiments}

\subsection{Data}
The effectiveness of our class combination approaches is evaluated on the organic dataset   \cite{b1}. It is gathered to explore organic food beliefs from consumers. The beliefs are annotated by a market researcher on online comments posted on forums and discussion boards of news websites from both English and German-speaking countries. These beliefs are further structured manually into superordinate themes and main themes. The dataset is multi-labeled, since numerous comments have multiple beliefs according to the research. We choose specifically a domain-specific dataset to research how our approach is able to fulfill the requirement even when there is a lack of data. The statistics of the datasets are summarized in Table \ref{tab1}. 

% \begin{table}[t]
% \caption{Statistics of the organic dataset}
% \begin{center}
% \setlength\tabcolsep{3.1pt}
% \begin{tabular}{lrrrrr}
% Language         & Main themes & Themes & Beliefs & Documents & Sentences \\ \hline
% English  & 4           & 21     & 62      & 1099      & 2275      \\
% German  & 4           & 21     & 60      & 789       & 2334     
% \end{tabular}

% \label{tab1}
% \end{center}
% \end{table}

\begin{table}[t]
\centering
\begin{tabular}{lll}
\hline
    Language & English & German \\ \hline
    Main themes & 4 & 4 \\
    Themes & 21 & 21 \\
    Beliefs & 62 & 60 \\
    Documents & 1099 & 789 \\
    Sentences & 2275 & 2334 \\ \hline
\end{tabular}
\caption{Statistics of the organic dataset}
\label{tab1}
\end{table}

\paragraph{Preprocessing}

Preprocessing consists of 5 steps. Sentence segmentation extracts sentences from documents which will be used to create sentence-level embeddings for unsupervised approaches. Text cleaning utilizes libraries and regular expressions to filter out URLs, stop words, brackets, quotes, line feeds and blank symbols. 
%Multi-label documents are merged using aggregate and group-by functions in Pandas. 
Optimal number of clusters are computed using decision criteria such as AIC, BIC scores together with Elbow method. 
%Fine-tuning of GloVe utilizes Mittens package   \cite{b2} to fine-tune the GloVe word embedding on the organic dataset. We aimed to research if the domain-specific fine-tuned word embedding would boost the performance of our embedding-based class combination approaches, i.e., word movers' distance and CF-IDF.

\subsection{Baseline: Expert-Based Class Labels}

We take the manually annotated labels and their class hierarchy as given by the domain expert as a baseline and compare them with the class hierarchies obtained from our hierarchical class clustering approach regarding classification performance. The manually annotated label consists of 4 main themes and 21 superordinate themes. Accordingly, we use hierarchical class clustering to generate 4 coarse classes and 21 fine-grained classes.

% \begin{figure*}[t]
%     \centering
%     \includegraphics[width=1\textwidth]{overview.png}
%     \caption{}
% \end{figure*}

\subsection{Hierarchically Clustered Class Labels}

The idea of hierarchically clustered class labels is to analyse the intrinsic semantic class hierarchy of existing expert labels. This is achieved by pairwise combination of fine-grained classes of social media texts according to their overall semantic similarities. The latter is measured by applying LSI as a distance metric on varying document representations. We investigate how the degree of class combination relates to classification accuracy. 

%Further considered hyperparameters are the dimension of the latent space is chosen between 4, 8, 16, 21, 60, 62, and 100 corresponding to the size of expert labels and the result of the optimal number of clusters analysis. Besides, we investigate which word embedding technique (word2vec, GloVe, fasttext) delivers the best performance for our embedding-based class combination approaches

There are 2 main steps for supervised class combination approaches, namely semantic similarity extraction and class combination using agglomerative hierarchical clustering. Semantic similarity extraction is realized by calculating distances on document representations. Agglomerative hierarchical clustering combines similar classes into new abstract classes based on distance matrices obtained from semantic similarity extraction. The closest clusters indicated by the Euclidean distance will be successively merged until K (4 or 21) clusters are formed.

\begin{figure*}[t]
    \centering
    \includegraphics[width=1\textwidth]{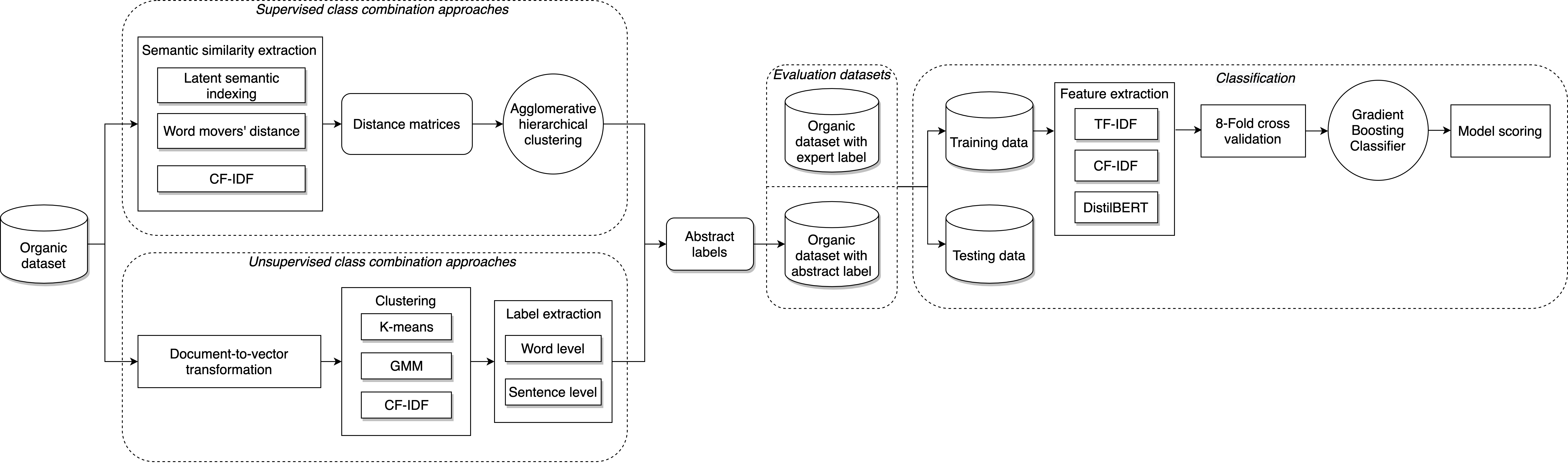}
    \caption{An overview of the experimental design. Abstract labels were extracted using supervised and unsupervised class combination approaches. They will be evaluated in a classification task competing with manually annotated expert label as baseline.}
\end{figure*}

\subsection{Classes from Unsupervised Clustering}

In contrast to the hierarchical clustering approach, which we apply on fine-grained \textit{expert classes} to generate more abstract classes, in the unsupervised approach, we assign labels automatically without incorporating any prior information. This is achieved by K-means clustering on multi-lingually aligned pre-trained textual embeddings. Therefore, we set the number of clusters to 4 or 21 corresponding to the number of main themes and superordinate themes from the expert class hierarchy. This allows us to make a comprehensive comparison of classification performance between labels obtained from the expert-labeled raw dataset, supervised class combination approach, and unsupervised class combination approach.

Each document is first segmented into sentences or words respectively. After clustering all according embeddings from the corpus, the cluster of each document is the most occurring cluster of all its respective sentences or words. A tie results in a multi-labeled document instance.

Since our organic datasets are bilingual, it is preferred to keep the consistency of word and sentence representations irrespective of the language being used. Thus, bilingually aligned representations of textual embeddings between English and German are preferred. 
Word embeddings are produced by multi-lingually aligned fasttext word vectors computed on Wikipedia \cite{b8}. Sentence embeddings are provided by the multilingual universal sentence encoder XLING \cite{b10,b11} trained with a dual-encoder that learns tied representations using translation-based bridge tasks \cite{b11}.

%At word level, after applying clustering algorithms on all word embeddings of the vocabulary, each word belongs to one cluster. Since a document normally consists of more than one word, each document can be regarded as a combination of word-level clusters. We use max-pooling which calculates the maximum word-level label occurrences of the document to determine its abstract label. If there are equal word-level label occurrences, this document would have a multi-label.

\subsection{Classification}

We evaluate the performance of our class combination approach on the organic dataset using the gradient boosted decision trees classifier based on tf-idf document features. The dataset can be categorized into two document classification tasks, namely 4-labelled classification, and 21-labelled classification. We use 80\% of the data for training and 20\% for testing. Besides, we apply 8-fold cross-validation, which divides the data into 8 folds and ensures that each fold is used as testing set in evaluation. The mean value of the scores from each iteration of 8-fold cross validation determines the overall performance of the model.

Since all classification tasks in our experiment are multi-label, exact match, F1 macro, F1 micro, and normalized entropy are the metrics that are used to measure the model performance. Exact match indicates the percentage of samples that have all their labels classified correctly. 
%Hamming loss is the fraction of the wrong labels to the total number of labels. Since it is a loss function, its optimal value is zero and its upper bound is one. 
F1 score is the harmonic mean of precision and recall. It reaches its best value at 1 and worst value at 0. F1 micro and F1 macro differ from each other on the type of averaging performed on the data. F1 micro calculates metrics globally by counting the total true positives, false negatives, and false positives. On the other hand, F1 macro calculates metrics for each class separately and finds their unweighted mean \cite{b14}. Information Entropy, amongst others, is a measure of disorder and uncertainty \cite{b15}, which we use to depict how balanced the classes are with respect to the number of their containing class samples. Normalization makes the distribution of entropy between 0 and 1, where 1 indicates perfectly balanced classes and 0 maximally unbalanced classes. Using these measures, we aim to research what are the favorable and unfavorable effects of class abstraction and how do these relate to each other.

%Besides the label difference, we research how do document representation techniques affect the classification by comparing the classification performance using 3 different document representations as feature vectors, namely tf-idf, cf-idf, and Transformers DistilBERT. Document's tf-idf vector representation is a vocabulary-sized tf-idf score matrix. Document's cf-idf vector representation is a concept-sized cf-idf score matrix, where concept size is set default to 100. Document's DistilBERT vector representation is a 768-dimensional contextual embedding based on the [CLS] token from Transformers  \cite{b16}. 

\section{Results}
\label{sec:results}

We summarize the main results in Table \ref{summary}, which shows that the classification results based on hierarchically clustered and unsupervised class labels overall outperform the ones given by the domain expert baseline. The performance gain is especially obvious for hierarchically clustered class labels in exact matches and f1 micro, while for unsupervised classes in F1 macro and normalized entropy. In the remainder of this section, we discuss the results of classification regarding the previous research questions in detail.

\begin{table*}[t]
\centering
\begin{tabular}{cccccccccccccc}
\hline
\multicolumn{1}{c|}{\multirow{2}{*}{Corpus}} & \multicolumn{1}{l|}{\multirow{2}{*}{Labels}} & \multicolumn{4}{c|}{Expert CA} & \multicolumn{4}{c|}{Supervised CA} & \multicolumn{4}{c}{Unsupervised CA} \\
\multicolumn{1}{c|}{} & \multicolumn{1}{l|}{} & \multicolumn{1}{l}{Em} &  \multicolumn{1}{l}{Fma} & \multicolumn{1}{l}{Fmi} & \multicolumn{1}{l|}{Ne} & \multicolumn{1}{l}{Em} & \multicolumn{1}{l}{Fma} & \multicolumn{1}{l}{Fmi} & \multicolumn{1}{l|}{Ne} & \multicolumn{1}{l}{Em} &  \multicolumn{1}{l}{Fma} & \multicolumn{1}{l}{Fmi} & \multicolumn{1}{l}{Ne} \\ \hline
\multirow{2}{*}{EN} & 4 & 0.39 & 0.48 & 0.54 & 0.92 & 
\textbf{0.66} & 0.44 & \textbf{0.76} & 0.63 &
%\textbf{0.67} & 0.53 & \textbf{0.76} & 0.62 & 
0.45 & \textbf{0.54} & 0.60 & \textbf{0.98} \\
& 21 & 0.17 & 0.22 & \textbf{0.33} & 0.85 & 
0.14 & 0.18 & 0.27 & 0.86 &
%\textbf{0.19} & 0.18 & 0.32 & 0.81 & 
\textbf{0.18} & \textbf{0.28} & 0.31 & \textbf{0.97} \\ \hline
\multirow{2}{*}{DE} & 4 & 0.25 & 0.37 & 0.43 & 0.92 &
\textbf{0.74} & 0.34 & \textbf{0.80} & 0.43 &
%\textbf{0.75} & 0.31 & 0.81 & 0.46 & 
0.32 & \textbf{0.61} & 0.70 & \textbf{0.94} \\
& 21 & 0.16 & 0.13 & 0.27 & 0.86 & 
\textbf{0.17} & 0.13 & 0.31 & 0.73 &
%\textbf{0.21} & 0.13 & \textbf{0.37} & 0.72 & 
0.08 & \textbf{0.31} & \textbf{0.35} & \textbf{0.98}
\end{tabular}
\caption{Exact match (Em), F1 macro (Fma), F1 micro (Fmi) , normalized entropy (Ne) for manual domain-expert class abstraction (baseline), best variants of supervised automatic class abstraction (CA) and best variants of unsupervised automatic class abstraction on 4-labelled and 21-labelled English and German text classification tasks based on TF-IDF document representation.}
\label{summary}
\end{table*}

\subsection{Coarse vs. Fine-Grained Classes}
%\label{degree}

We compare how the number of classes, i.e., few classes with many samples versus many classes with few samples, would relate to classification accuracy against the background of different class hierarchies and methodologies. We consider 4 and 21 as possible number of classes and respective clusters for the unsupervised case.

Regarding macro F1 scores, these appear the best for unsupervised class labels followed by expert class labels. There is the same relation regarding entropy, i.e., there is a clear relation with regards to how well the classes are actually balanced. Micro F1 scores and exact matches tend to better better for hierarchically clustered classes at times where classes are rather imbalanced. From these observations it appears that the issue of balanced classes might be related to how a classifier is going to perform.

% \begin{figure}[htbp]
% \centerline{\includegraphics[width=0.5\textwidth]{Radar_degree.png}}
% \caption{Classification performance regarding degree of class abstraction. Supervised LSI-based and CF-IDF-based class abstractions with different latent space dimensionalities were used to generate abstract labels. Those abstract labels were evaluated in 4-labelled and 21-labelled organic English text classification tasks.}
% \label{degree}
% \end{figure}

%Based on Fig. \ref{degree}, degree of class abstraction has more impact on LSI-based class abstraction than CF-IDF-based class abstraction, especially for 4-labelled classification tasks. Despite dissimilar degrees of class abstraction, CF-IDF-based class abstraction still yields similar results. It is worth point out that the degree 4 and 21 corresponding to the label sizes provide most of the cases the maximum entropy in classification. Besides, they often perform as well as the other degrees in the remaining metrics. Thus, based on our experiment it is recommendable to choose a label-sized degree as the desired degree for abstraction. 

\subsection{Side Effects}
\label{effects}

We observe in table \ref{summary} that although our hierarchical class clustering approach provides high exact match and F1 micro scores, it also suffers from extremely low F1 macro and normalized entropy scores. From both it can be concluded that this is due to an imbalanced class distribution. It results from those fine-grained expert classes which have a) a low number of documents and b) are semantically highly dissimilar to the other existing classes. This leads to combined classes and regarding class hierarchies containing a comparatively small number of samples. These cannot represent a given class sufficiently for a machine learning classifier. It can be considered as an undesired side effect of hierarchical class clustering, which in its algorithm does not consider class balancing as opposed to k-means for example. As the approach is completely based on existing expert classes which sometimes are unbalanced and dissimilar, this outcome is inevitable. 
%Despite those obstacles, we still tried to achieve better accuracy with as much higher entropy as possible. 
Thus, class imbalance is an unfavourable effect of guided class abstraction.

The expert-based class structure with its 4 and 21 classes, respectively, is more balanced. This means that the way in which the domain expert structures the class hierarchies appears to be more beneficial with respect to class balancing and thus classification with respect to macro F1 scores. However, micro F1 scores are generally small here, which raises questions about how well classification overall might work.

On the other hand, the unsupervised clustering approach does not have these issues and achieves a satisfying result. Its F1 micro, F1 macro, and normalized entropy exceed domain-expert class abstraction by significant margin, while exact match is almost identical. Besides, it can discover new interesting fine-grained classes such as milk that maybe ignored by domain experts. This result shows the beneficial properties of the approach for effective classification, and it demonstrates that it could improve classification over manual, domain expert-based class labels.

\section{Discussion}
\label{sec:conclusion}

%\subsection{Findings}

We presented a systematic analysis of 3 class abstraction approaches on the organic datasets for text classification tasks. In accordance with the previously explained results, we answer the research questions from Section \nameref{introduction} as follows:%\\

\paragraph{RQ1: \rqone} 
%\hfill\\ %\hfill\\ 
Table \ref{summary} summarizes the classification accuracy of all class abstraction approaches. It can be seen that the domain expert classes offer moderate performances. The macro F1 scores lie between the ones from the hierarchical and unsupervised class labels, which is supported by the fact that the classes tend to be balanced well according to the entropy. However, the micro F1 scores and exact matches are mostly the lowest among all approaches. 

%The potential reason behind it could be the expert class outliers explained in Subsection \nameref{effects}.
%\\

\paragraph{RQ2: \rqtwo} 
%\hfill\\ %\hfill\\ 
According to our measurements, a fewer number of classes generally improves classification performance significantly, independent of how the class labels are derived, i.e., by an expert, hierarchical class clustering, or unsupervised clustering. By reducing the number of classes from 21 to 4, macro and micro F1 scores at least doubled for all methods. The improvements are specifically striking for hierarchical class clustering. This means that for classification on small, expert-labeled opinion mining datasets with fine-grained labels, a reduction of the number of classes should be considered as an option.

%With the results in Fig. \ref{degree}, we can conclude that degree of class abstraction has more impact on LSI-based than CF-IDF-based class abstraction approaches. We discover that a label-sized degree would be the optimal degree for abstraction. This finding not only minimizes the uncertainty, but also simplifies the research process, and reduces extra fine-tuning efforts. 
%\\

\paragraph{RQ3: \rqthree} 
%\hfill\\ \hfill\\ 
The classification results in table \ref{summary} show that our hierarchical class clustering approach improves micro F1 classification performance over manual, domain expert-based class labels. However, class imbalancing is a problematic side effect, which can lead to low macro F1 scores. Further research needs to determine if and how this issue can be solved on the given problem domain.
%This analysis provides an example to the market researchers on how to leverage artificial intelligence or rather machine learning and NLP techniques to perform large-scale automated content analysis such as building a category system even without considerable expertise in a domain.
%\\

\paragraph{RQ4: \rqfour} 
%\hfill\\ \hfill\\ 
We observe in Table \ref{summary} that our unsupervised clustering approach outperforms expert-based classes with margin for all performance metrics. In the related work, it has been shown that this approach is also able to generate a meaningful intrinsic class hierarchy based on the available data without any human interference \cite{dannerMediaAgenda,gerryMediaAgenda}. Our finding shows that automated class abstraction can also serve serve as an alternative to domain-expert class labeling for classification. We perceive a great potential for it and hope the proposed technique would offer valuable guidance for market researchers investigating social media.
%\\

\paragraph{RQ5: \rqfive} 
%\hfill\\ \hfill\\ 
As explained in Subsection \nameref{effects}, hierarchical class clustering suffers from minority classes, which is introduced by small outlier classes of the expert annotations. Those classes which have a low amount of documents are combined and are then not semantically representative enough in classification. Consequentially, they lead to imbalanced abstract class distributions. Thus, achieving not only a satisfying precision but also a high entropy is a demanding task for hierarchical class clustering. Nevertheless, our unsupervised class abstraction approach has overcome this issue and clearly improves over the expert based classes baseline.

% \subsubsection{RQ6 \rqsix} \hfill\\
% \hfill\\ As shown in Table \ref{embedding}, sentence-level unsupervised class abstraction is proven to be more efficient than word-level. As possible reasons, we recognize label extraction on word-level to be an extremely challenging task as applying simple max-pooling may lead to over-redundant abstract labels.\\

% \subsubsection{RQ7 \rqseven} \hfill\\
% \hfill\\ Given the classification accuracy, according to Fig. \ref{d_document} and Fig. \ref{k_document}, we discover tf-idf document representation combines well with non-embedding-based class abstraction, while cf-idf document representation is particularly suited for embedding-based class abstraction.

\section{Conclusion}

The described experiments and results show that it is technically feasible to classify belief statements automatically. However, the fine-grained nature of labels given by a domain expert make the task challenging. Combining classes of belief statements is helpful, as it makes the data statistically more representative for machine learning algorithms. Automatized class combinations might lead to imbalanced classes, but overall better classification scores. Classes given by our proposed unsupervised approach yield very balanced classes, which give best classification results. We conclude by recommending our unsupervised clustering approach based on deep, multi-lingual, and pre-trained sentence embeddings, which showed beneficial results in previous opinion mining studies, too \cite{dannerMediaAgenda,gerryMediaAgenda}.

\bibliographystyle{acl_natbib}
\bibliography{main}

\begin{thebibliography}{18}
\expandafter\ifx\csname natexlab\endcsname\relax\def\natexlab#1{#1}\fi

\bibitem[{Bakharia(2019)}]{10.1007/978-3-030-33232-7_25}
Aneesha Bakharia. 2019.
\newblock On the equivalence of inductive content analysis and topic modeling.
\newblock In \emph{Advances in Quantitative Ethnography}, pages 291--298, Cham.
  Springer International Publishing.

\bibitem[{Bojanowski et~al.(2017)Bojanowski, Grave, Joulin, and Mikolov}]{b8}
Piotr Bojanowski, Edouard Grave, Armand Joulin, and Tomas Mikolov. 2017.
\newblock Enriching word vectors with subword information.
\newblock \emph{Transactions of the Association for Computational Linguistics},
  5:135--146.

\bibitem[{Cer et~al.(2018)Cer, Yang, Kong, Hua, Limtiaco, John, Constant,
  Guajardo-C{\'e}spedes, Yuan, Tar et~al.}]{b10}
Daniel Cer, Yinfei Yang, Sheng-yi Kong, Nan Hua, Nicole Limtiaco, Rhomni~St
  John, Noah Constant, Mario Guajardo-C{\'e}spedes, Steve Yuan, Chris Tar,
  et~al. 2018.
\newblock Universal sentence encoder.
\newblock \emph{arXiv preprint arXiv:1803.11175}.

\bibitem[{Chidambaram et~al.(2018)Chidambaram, Yang, Cer, Yuan, Sung, Strope,
  and Kurzweil}]{b11}
Muthuraman Chidambaram, Yinfei Yang, Daniel Cer, Steve Yuan, Yun-Hsuan Sung,
  Brian Strope, and Ray Kurzweil. 2018.
\newblock Learning cross-lingual sentence representations via a multi-task
  dual-encoder model.
\newblock \emph{arXiv preprint arXiv:1810.12836}.

\bibitem[{Danner et~al.(N.D.)Danner, Hagerer, Pan, and
  Groh}]{dannerMediaAgenda}
Hannah Danner, Gerhard Hagerer, Yan Pan, and Georg Groh. N.D.
\newblock The news media and its audience: Agenda-setting on organic food in
  the united states and germany.
\newblock Currently under review.

\bibitem[{Danner and Menapace(2020)}]{b1}
Hannah Danner and Luisa Menapace. 2020.
\newblock Using online comments to explore consumer beliefs regarding organic
  food in german-speaking countries and the united states.
\newblock \emph{Food Quality and Preference}, 83:103912.

\bibitem[{Deerwester et~al.(1990)Deerwester, Dumais, Furnas, Landauer, and
  Harshman}]{b3}
Scott Deerwester, Susan~T Dumais, George~W Furnas, Thomas~K Landauer, and
  Richard Harshman. 1990.
\newblock Indexing by latent semantic analysis.
\newblock \emph{Journal of the American society for information science},
  41(6):391--407.

\bibitem[{Hagerer et~al.(N.D.)Hagerer, Leung, Danner, and
  Groh}]{gerryMediaAgenda}
Gerhard Hagerer, Wing~Sheung Leung, Hannah Danner, and Georg Groh. N.D.
\newblock A case study and qualitative analysis of simple cross-lingual opinion
  mining.
\newblock Currently under review.

\bibitem[{He et~al.(2013)He, Zha, and Li}]{HE2013464}
Wu~He, Shenghua Zha, and Ling Li. 2013.
\newblock \href
  {https://doi.org/https://doi.org/10.1016/j.ijinfomgt.2013.01.001} {Social
  media competitive analysis and text mining: A case study in the pizza
  industry}.
\newblock \emph{International Journal of Information Management},
  33(3):464--472.

\bibitem[{Lin(1991)}]{b15}
Jianhua Lin. 1991.
\newblock Divergence measures based on the shannon entropy.
\newblock \emph{IEEE Transactions on Information theory}, 37(1):145--151.

\bibitem[{{Maimon} and {Rokach}(2005)}]{b6}
Oded {Maimon} and Lior {Rokach}. 2005.
\newblock \emph{Data Mining and Knowledge Discovery Handbook}.
\newblock Springer.

\bibitem[{Martin and Turner(1986)}]{martin1986grounded}
Patricia~Yancey Martin and Barry~A Turner. 1986.
\newblock Grounded theory and organizational research.
\newblock \emph{The journal of applied behavioral science}, 22(2):141--157.

\bibitem[{Pedregosa et~al.(2011)Pedregosa, Varoquaux, Gramfort, Michel,
  Thirion, Grisel, Blondel, Prettenhofer, Weiss, Dubourg et~al.}]{b14}
Fabian Pedregosa, Ga{\"e}l Varoquaux, Alexandre Gramfort, Vincent Michel,
  Bertrand Thirion, Olivier Grisel, Mathieu Blondel, Peter Prettenhofer, Ron
  Weiss, Vincent Dubourg, et~al. 2011.
\newblock Scikit-learn: Machine learning in python.
\newblock \emph{the Journal of machine Learning research}, 12:2825--2830.

\bibitem[{Perner(2018)}]{pernerconsumer}
Lars Perner. 2018.
\newblock \href {https://www.consumerpsychologist.com/cb_Attitudes.html}
  {Consumer behavior - attitudes}.
\newblock [Online; accessed 29-June-2021].

\bibitem[{Piepenbrink and Gaur(2017)}]{piepenbrink2017topic}
Anke Piepenbrink and Ajai~Singh Gaur. 2017.
\newblock Topic models as a novel approach to identify themes in content
  analysis.
\newblock In \emph{Academy of Management Proceedings}, volume 2017, page 11335.
  Academy of Management Briarcliff Manor, NY 10510.

\bibitem[{Rocklage and Rucker(2019)}]{rocklage2019text}
Matthew~D Rocklage and Derek~D Rucker. 2019.
\newblock Text analysis in consumer research: An overview and tutorial.
\newblock \emph{Handbook of Research Methods in Consumer Psychology}, pages
  385--402.

\bibitem[{Xu et~al.(2011)Xu, Liao, Li, and Song}]{XU2011743}
Kaiquan Xu, Stephen~Shaoyi Liao, Jiexun Li, and Yuxia Song. 2011.
\newblock \href {https://doi.org/https://doi.org/10.1016/j.dss.2010.08.021}
  {Mining comparative opinions from customer reviews for competitive
  intelligence}.
\newblock \emph{Decision Support Systems}, 50(4):743--754.
\newblock Enterprise Risk and Security Management: Data, Text and Web Mining.

\bibitem[{Yu et~al.(2011)Yu, Jannasch-Pennell, and
  DiGangi}]{yu2011compatibility}
Chong~Ho Yu, Angel Jannasch-Pennell, and Samuel DiGangi. 2011.
\newblock Compatibility between text mining and qualitative research in the
  perspectives of grounded theory, content analysis, and reliability.
\newblock \emph{Qualitative Report}, 16(3):730--744.

\end{thebibliography}

\end{document}